\documentclass{article}

\usepackage{microtype}
\usepackage{graphicx}
\usepackage{subfigure}
\usepackage{booktabs} 

\usepackage{hyperref}

\usepackage{tikz}
\newcommand*\circled[1]{\tikz[baseline=(char.base)]{
            \node[shape=circle,draw,inner sep=1.5pt] (char) {#1};}}

\usepackage[accepted]{icml2022}

\usepackage{amsmath}
\usepackage{amssymb}
\usepackage{mathtools}
\usepackage{amsthm}
\DeclareMathAlphabet\mathbfcal{OMS}{cmsy}{b}{n}

\newcommand{\ten}[1]{\mathbfcal{#1}} 
\newcommand{\mat}[1]{\mathbf{#1}}

\usepackage[capitalize,noabbrev]{cleveref}

\theoremstyle{plain}

\theoremstyle{definition}

\theoremstyle{remark}

\usepackage[textsize=tiny]{todonotes}

\icmltitlerunning{TT-PINN: A Tensor-Compressed Neural PDE Solver for Edge Computing}

\begin{document}

\twocolumn[
\icmltitle{TT-PINN: A Tensor-Compressed Neural PDE Solver for Edge Computing}



\icmlsetsymbol{equal}{*}

\begin{icmlauthorlist}
\icmlauthor{Ziyue Liu}{equal,pstat}
\icmlauthor{Xinling Yu}{equal,ece}
\icmlauthor{Zheng Zhang}{ece}
\end{icmlauthorlist}

\icmlaffiliation{pstat}{Department of Statistics and Applied Probability, University of California, Santa Barbara, CA, United States}
\icmlaffiliation{ece}{Department of Electrical and Computer Engineering, University of California, Santa Barbara, CA, United States}

\icmlcorrespondingauthor{Ziyue Liu}{ziyueliu@ucsb.edu}
\icmlcorrespondingauthor{Xinling Yu}{xyu644@ucsb.edu}
\icmlcorrespondingauthor{Zheng Zhang}{zhengzhang@ece.ucsb.edu}

\icmlkeywords{Machine Learning, ICML}

\vskip 0.3in
]



\printAffiliationsAndNotice{\icmlEqualContribution} 

\begin{abstract}
Physics-informed neural networks (PINNs) have been increasingly employed due to their capability of modeling complex physics systems. To achieve better expressiveness, increasingly larger network sizes are required in many problems. This has caused challenges when we need to train PINNs on edge devices with limited memory, computing and energy resources. To enable training PINNs on edge devices, this paper proposes an end-to-end compressed PINN based on Tensor-Train decomposition. In solving a Helmholtz equation, our proposed model significantly outperforms the original PINNs with few parameters and achieves satisfactory prediction with up to 15$\times$ overall parameter reduction.
\end{abstract}

\section{Introduction}
Physics-informed neural networks (PINNs) are increasingly used to solve a wide range of forward and inverse problems involving partial differential equations (PDEs), including fluids mechanics \cite{raissi2020hidden}, materials modeling \cite{liu2019multi}, safety verification \cite{bansal2021deepreach} and control \cite{onken2021neural} of autonomous systems. Despite their success of learning complex systems using the simple multilayer perception (MLP) architecture, large neural networks are often required to achieve high expressive power. This has significantly increased the memory and computing cost of training a PINN. Furthermore, a PINN often has to be trained many times in practice once the problem setting (e.g., boundary condition, measurement data, safety specification) changes. 

It is increasingly important to enable PINN training on resource-constraint edge devices. On one side, safety-aware learning-based verification and control~\cite{bansal2021deepreach,onken2021neural} often require the PINN to be trained on a tiny embedded processor of an autonomous agent. On the other side, the emerging digital twin and smart manufacturing need AI-assistant design with IP protection~\cite{stevens2020ai}, where federated learning with many edge devices allows users to design shared AI models without disclosing their private data. In both cases, training has to be done on edge devices with very limited memory, computing and energy budget. 

This paper proposes {\bf TT-PINN}, an end-to-end tensor-compressed method for training PINNs. This method achieves huge parameter and memory reduction in the training process, by combining Tensor-Train compressed model representation and a physics-informed network to approximate the solutions of PDEs.
We use this method to solve a Helmholtz equation and compare it with standard PINNs. With only thousands of parameters, our models significantly outperform the original PINNs of similar or larger sizes. 

\section{Background: PINN}
We consider the problem of solving a PDE 
\begin{equation}
\begin{aligned}
&\boldsymbol{u}_{t}+\mathcal{N}_{\boldsymbol{x}}[\boldsymbol{u}]=0, \quad \boldsymbol{x} \in \Omega, t \in[0, T] \\
&\boldsymbol{u}(\boldsymbol{x}, 0)=h(\boldsymbol{x}), \quad x \in \Omega, \\
&\boldsymbol{u}(\boldsymbol{x}, t)=g(\boldsymbol{x}, t), \quad t \in[0, T], \quad \boldsymbol{x} \in \partial \Omega
\end{aligned}
\label{PDE}
\end{equation}
where $\boldsymbol{x}$ and $t$ are the spatial and temporal coordinates respectively, $\Omega$ and $\partial \Omega$ denote the computational domain and its boundary; $\mathcal{N}_{\boldsymbol{x}}$ is a general linear or nonlinear operator; $\boldsymbol{u}(\boldsymbol{x}, t)$ is the solution of the above PDE with the initial condition $h(\boldsymbol{x})$ and the boundary condition. In PINNs~\cite{raissi2019physics}, a neural network approximation  $\boldsymbol{u}(\boldsymbol{x}, t) \approx f_{\boldsymbol{\theta}}(\boldsymbol{x}, t)$ parameterized by $\boldsymbol{\theta}$ is substituted into the PDE \eqref{PDE} and yields a residual defined as
\begin{equation}
\boldsymbol{r}_{\boldsymbol{\theta}}(\boldsymbol{x}, t):=\frac{\partial}{\partial t} f_{\boldsymbol{\theta}}(\boldsymbol{x}, t)+\mathcal{N}_{\boldsymbol{x}}\left[f_{\boldsymbol{\theta}}(\boldsymbol{x}, t)\right].
\end{equation} 
We train parameters $\boldsymbol{\theta}$ by minimizing the loss function
\begin{equation}
\mathcal{L}=\mathcal{L}_{r}+\mathcal{L}_{b}+\mathcal{L}_{0}.
\end{equation}
Here
\begin{equation}
\small
\begin{aligned}
&\mathcal{L}_{r}=\frac{1}{N_{r}} \sum_{i=1}^{N_{r}}\left|\boldsymbol{r}_{\boldsymbol{\theta}}(\boldsymbol{x_{r}^{i}}, t_{r}^{i})\right|^{2}, \mathcal{L}_{b}=\frac{1}{N_{b}} \sum_{i=1}^{N_{b}}\left|f_{\boldsymbol{\theta}}\left(\mathbf{x}_{b}^{i}, t_{b}^{i}\right)-g_{b}^{i})\right|^{2}, \\
&\mathcal{L}_{0}=\frac{1}{N_{0}} \sum_{i=1}^{N_{0}}\left|f_{\boldsymbol{\theta}}\left(\mathbf{x}_{0}^{i}, 0\right)-h_{0}^{i})\right|^{2}
\end{aligned}
\normalsize
\end{equation}
penalize the residual of the PDE, the boundary conditions and the initial conditions respectively; $N_{r}$, $N_{b}$, and $N_{0}$ are the numbers of data points for corresponding loss terms.

\begin{figure}[t]
\vskip 0in
\begin{center}
\centerline{\includegraphics[width=1\columnwidth]{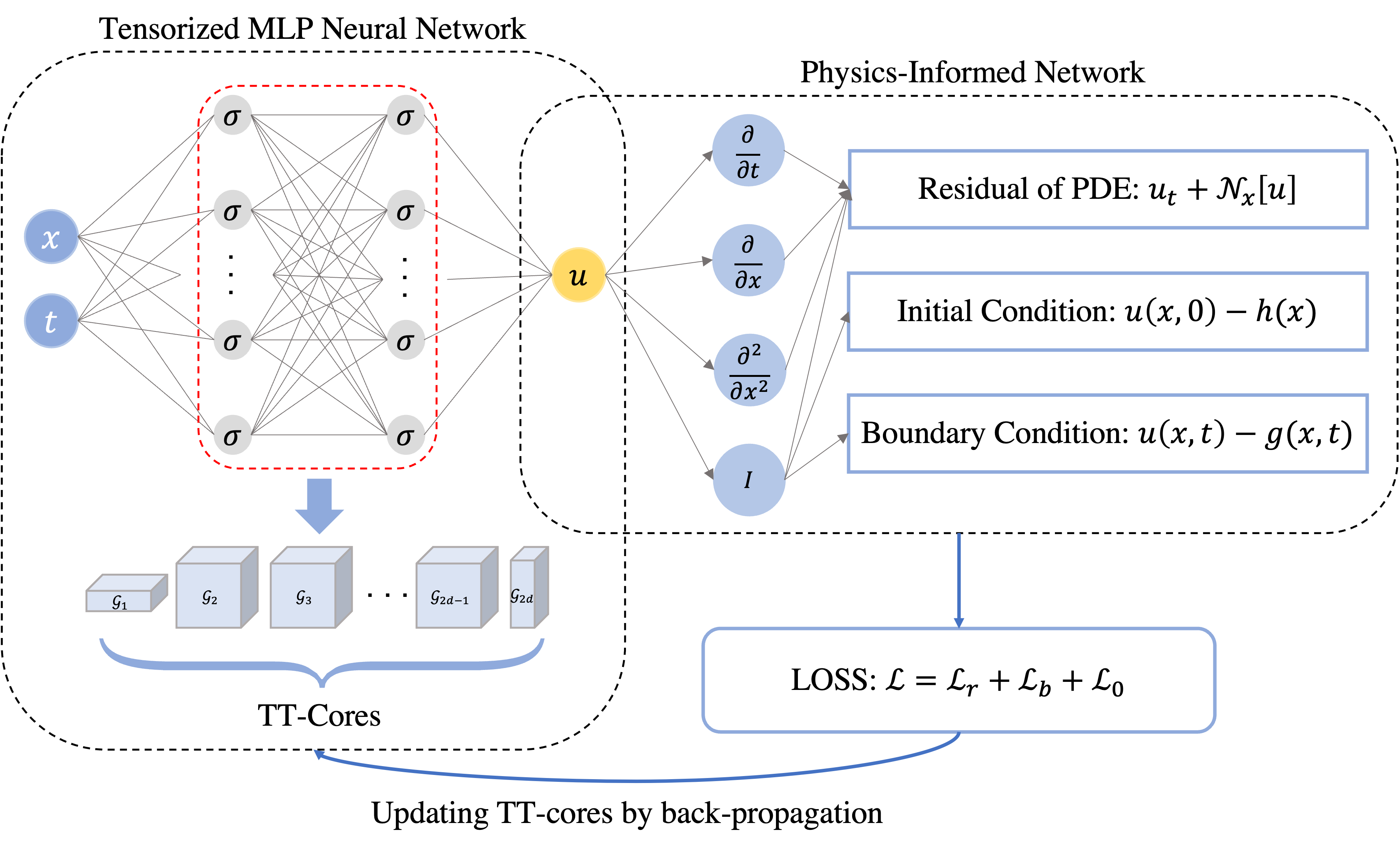}}
\caption{The proposed TT-PINN framework. The Left part is a tensorized MLP neural network with one tensorized hidden layer, in which the trainable parameters are stored as multiple TT-cores. The right part is the embedded governing physical laws that are used to identify the PDE and to force the neural network approximating the solution. During the training, the TT-cores are directly updated by its gradients calculated via auto-differentiation.}
\label{fig1}
\end{center}
\vskip -0.4in
\end{figure}

\section{The TT-PINN Method}

\subsection{TT-PINN Architecture} 
In this work, we consider tensor-compressed training of PINN based on a multilayer perception (MLP) network. A standard MLP uses an $L$-layer cascaded function
\begin{equation}
\label{fwd}
\boldsymbol{z}_k=\sigma\left(\mat{W}_k \boldsymbol{z}_{k-1} + \boldsymbol{b}_k\right), \; k=1, 2, \cdots, L
\end{equation}
with $\boldsymbol{z}_0=[\boldsymbol{x},t]$ and $ \boldsymbol{u}=\boldsymbol{z}_L$ to approximate the solution. The weight matrix $\mat{W}_k$ can consume lots of memory, making the training unaffordable on edge devices. This challenge becomes more significant when the PDE operator involves highly inhomogeneous material properties or strongly scattered waves. In these cases,  large neural networks are often needed to obtain high expressive powers. 

As shown in the Figure \ref{fig1}, the TT-PINN replaces the weight matrix of an MLP layer by a series of TT-cores in the training process. For simplicity, we drop the layer index, and let $\mathbf{W} \in \mathbb{R}^{M\times N}$ denote a generic weight matrix in an MLP layer. We factorize its dimension sizes as $M = \prod^{d} \limits_{i=1}m_i$ and $N = \prod^{d} \limits_{j=1}n_j$, fold $\mathbf{W}$ into a $2d$-way tensor $\mathbfcal{W} \in \mathbb{R}^{m_1\times m_2 \times \dots \times m_d \times n_1 \times n_2 \times \dots \times n_d}$, and approximate $\ten{W}$ with the TT-decomposition~\cite{oseledets2011tensor}:
\begin{equation}
\begin{aligned}
&\widehat{\mathbfcal{W}}(i_1, i_2, \dots, i_d, j_1, j_2, \dots, j_d) \\
&= \mat{G}_1(i_1)\dots \mat{G}_d(i_d) \mat{G}_{d+1}(j_1)\dots \mat{G}_{2d}(j_d).
\end{aligned}
\end{equation}
Here $\mat{G}_k(i_k) \in \mathbb{R}^{r_{k-1} \times r_{k}}$ is the $i_k$-th slice of the TT-core $\mathbfcal{G}_k \in \mathbb{R}^{r_{k-1}\times m_k \times r_k}$by fixing its $2$nd index as $i_k$. The vector $(r_0, r_1, \dots, r_{2d})$ is called TT-ranks with the constraint $r_0 = r_{2d} = 1$. This TT representation reduces the number of unknown variables in a weight matrix from $\prod^{d} \limits_{k=1}m_k n_k$ to $\sum \limits_{k=1}^{d}r_{k-1}m_k r_k +r_{d+k-1}n_k r_{d+1}$. The compression ratio can be controlled by the TT-ranks. Recent approaches can learn proper TT-ranks automatically in the training process via a Bayesian formulation~\cite{hawkins2021bayesian,hawkins2022towards}.

In most existing works of TT-layer~\cite{novikov2015tensorizing}, a Tensor-Train-Matrix (TTM) decomposition is used, in which the weight matrix is represented by $d$ $4$-way TT-cores instead of $2d$ $3$-way TT-cores as we described above. Here we adopt TT instead of TTM, because the TT format allows easy tensor-network contraction (shown in Section~\ref{subsec:fwdbwd}), which can greatly reduce the memory and computational cost in both forward and backward propagation. 

\subsection{Forward \& Backward Propagation of TT-PINN}
\label{subsec:fwdbwd}

\begin{figure*}[t]
\begin{center}
\centerline{\includegraphics[width=2\columnwidth]{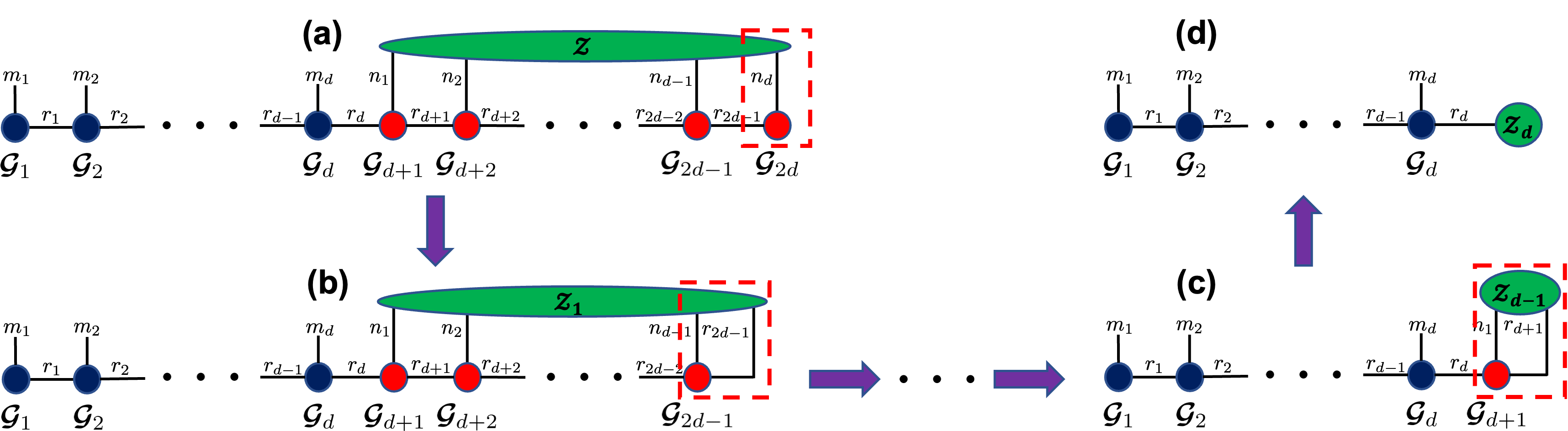}}
\caption{Matrix-vector product in the forward propagation using low-rank TT-cores only.}
\label{fig3}
\end{center}
\vskip -0.4in
\end{figure*}

Compared to techniques that compress a well-trained model for inference, TT-PINNs are directly trained in the compressed format. Specifically, the TT-cores that approximate a weight matrix are directly used in the forward propagation and updated in the backward propagation.

\paragraph{Memory-Efficient Forward Propagation.} As shown in \eqref{fwd}, the main cost in a forward pass is computing a matrix vector product like  $\mat{W}\boldsymbol{z}$. Instead of reconstructing $\mat{W}$ from its TT-cores, we directly use its low-rank TT-cores to obtain the result.  Specifically, let $\ten{Z}\in \mathbb{R}^{n_1 \times n_2 \cdots \times n_d}$ be the folding of $\boldsymbol{z}$ into a $d$-way tensor, then TT-PINN computes a series of tensor-network contractions between tensor $\ten{Z}$ and the TT-cores $\{\ten{G}_i\}_{i=1}^{2d}$ as shown in Fig.~\ref{fig3}. 

We use the tensor-network notation~\cite{orus2014practical,cichocki2014tensor} to show the computation process. A generic $N$-way tensor is represented by circle and $N$ edges; a shared edge among two tensors mean production (i.e., contraction) along that dimension. 
Before the computation starts, TT-cores are neither connected to each other nor connected to the tensor $\ten{Z}$. We now explain the whole process by three steps. \circled{1} Firstly, the tensor $\ten{Z}$ contracts with the last TT-core $\ten{G}_{2d}$ as shown by the red dashed rectangle in Fig. \ref{fig3} (a), producing an intermediate tensor $\ten{Z}_1$, in which the size of the $d$-th dimension changes from $n_d$ to $r_{2d-1}$ and all the other dimensions remains unchanged. \circled{2} In the second step, the rest of the red TT-cores are contracted in sequence, from $\ten{G}_{2d-1}$ to $\ten{G}_{d+1}$. Fig. \ref{fig3} (b) shows the contraction between the first intermediate tensor $\ten{Z}_1$ and $\ten{G}_{2d-1}$ on two dimensions, producing a $(d-1)$-way tensor $\ten{Z}_2 \in \mathbb{R}^{n_1 \times n_2 \times \cdots \times n_{d-2} \times r_{2d-2}}$. Similarly, each time the $k$-th intermediate tensor $\ten{Z}_k$ contracts with the $(2d-k)$-th TT-core $\ten{G}_{2d-k}$, and the resulting tensor $\ten{Z}_{k+1}$ will have one dimension eliminated. After $\ten{Z}_{d-1}$ contracts with the last red TT-core $\ten{G}_{d+1}$, the resulting tensor $\ten{Z}_{d}$ will only have one dimension of size $r_d$, as shown in Part (c) and (d) of Fig. \ref{fig3}. \circled{3} Finally, we contract $\ten{Z}_d$ with $\ten{G}_d$ and connect all the other TT-cores together by sequentially contracting $\ten{G}_d$ with $\ten{G}_{d-1}$, $\ten{G}_{d-2}$, and all the way to $\ten{G}_{1}$, obtaining the final result as a vector of size $m_1m_2\cdots m_d$.

\paragraph{Backward Propagation.} After the forward propagation, our proposed TT-PINNs calculate the customized loss function similarly as the traditional PINNs, then the backward propagation begins, in which the auto-differentiation (AD) algorithm~\cite{baydin2018automatic} is applied. Since the AD automatically records each computation step and the evolved objects during the forward pass to generate a so-called computational graph that is used to calculate the gradient of the loss w.r.t each object through the chain rule, we are able to obtain the gradient for each TT-core thus directly updating each TT-core using stochastic gradient descent.

Through the whole process of the forward and backward propagations in the proposed TT-PINNs, all computations are done on the compressed parameters, i.e., TT-cores, instead of a full-size weight matrix. Therefore, this end-to-end compressed training framework can largely reduce the memory cost during the training.

\section{Experiments and Results}

\begin{figure*}[t]
\begin{center}
\centerline{\includegraphics[width=2\columnwidth]{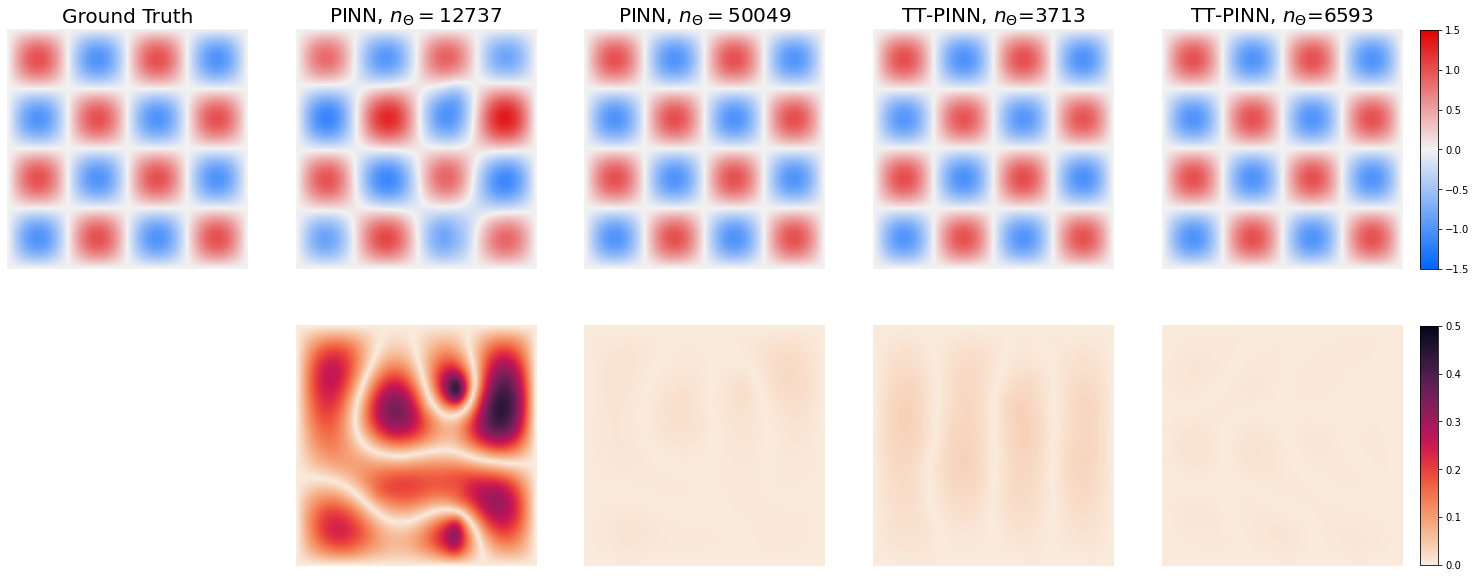}}
\caption{Comparison of PINNs and TT-PINNs in solving the Helmholtz equation \eqref{Helmholtz}. The first row contains the prediction results compared to the ground-truth solution. The second row shows the corresponding absolute errors. Our approach achieves similarly accurate prediction while using $15\times$ less parameters than the original PINN.}
\label{fig2}
\end{center}
\vskip -0.4in
\end{figure*}

In this section, we present a series of numerical studies to assess the performance of the proposed TT-PINN against a standard MLP PINN. Specifically, we consider a two-dimensional Helmholtz PDE: 
\begin{equation}
\begin{aligned}
     &  (\Delta+k^{2}) u(x, y)-g(x, y)=0, \quad(x, y) \in \Omega:=[0,1]^{2}, \\
     & u(x, y)=0, \quad(x, y) \in \partial \Omega,
\end{aligned}
\label{Helmholtz}
\end{equation}
where $\Delta$ is the Laplace operator and $k=4\pi $ is the wave number. The exact solution to this problem takes the form $u(x, y)=\sin \left(k x\right) \sin \left(k y\right)$, corresponding to a source term 
\begin{equation}
g(x, y)=k^{2} \sin \left(k x\right) \sin \left(k y\right).
\end{equation}

The PINN approximation $u_{\boldsymbol{\theta}}(x,y)$ to solving \eqref{Helmholtz} can be constructed by parametrizing its solution with a deep neural network $f_{\boldsymbol{\theta}}(x,y)$
\begin{equation}
u_{\boldsymbol{\theta}}(x,y)=x(x-1) y(y-1) f_{\boldsymbol{\theta}}(x,y).
\end{equation}
The above transformation is applied to the neural network to exactly meet the Dirichlet boundary condition \cite{lu2021physics}. Then the parameters $\boldsymbol{\theta}$ can be identified by
minimizing the total residual at
 $N_{r} = 1200$ collocation points that are randomly placed inside the domain $\Omega$. 

We use this benchmark problem to compare the performances of TT-PINNs against PINNs in terms of the total number of parameters. Specifically, we consider a set of neural networks with 3 hidden layers, and we control the number of parameters by varying the number of neurons per layer for PINNs and the choice of TT-ranks for TT-PINNs. In TT-PINNs, the TT-ranks were determined by the desired compression ratio for each hidden layer. For example, to compress a $ 256\times 256 $ weight matrix $\mat{W}_{h}$ in a fully-connected layer with $40\times$ compression, the TT-ranks are determined as $(1, 8, 8, 8, 8, 8, 8, 8, 1)$ when factorizing each dimension of $\mat{W}_{h}$ as $256=4^4$. It is also possible to automatically determine the TT-ranks via the Bayesian tensor rank determination in~\cite{hawkins2022towards,hawkins2021bayesian}. To guarantee convergence, all models are trained with 40,000 iterations. As for the training settings, we use the Adam optimizer \cite{kingma2014adam} with an initial learning rate $10^{-3}$ decayed by the factor of 0.9 after each 1000 iterations. The neural networks are initialized by the Xavier initialization scheme \cite{glorot2010understanding}, and a ${\rm Sine}$ activation function is applied to each neuron.

Table \ref{Tabel1} and Table \ref{Tabel2} summarize our results. Clearly, the expressive power of both standard PINNs and the proposed TT-PINN scales with its model size: larger models provide better approximation to the ground-truth solution. However, our proposed TT-PINNs achieves satisfactory prediction while using much less parameters than a fully connected 3-layer PINN with 256 neurons per layer. To avoid any confusion, the compression ratios reported in Table \ref{Tabel2} are for tensorized hidden layers, not for the whole model because so far we only tensorize the hidden layers and leave the input layer and the output layer uncompressed.

\begin{table}[t]
    
    \caption{The performance of PINNs in solving Helmholtz equation for different model sizes. Here $\mat{W}_{h}$ represents the weight matrix of each hidden layer and $n_{\boldsymbol{\theta}}$ is the total number of parameters. The mean squared errors and relative $\ell_2$ errors are reported.}
    \vskip 0.15in
    \centering
    \scalebox{0.9}{
    \begin{tabular}{cccc}
    \toprule  
    $\mat{W}_{h}$  & $n_{\boldsymbol{\theta}}$ & MSE & Rel. $\ell_{2} $ error \\
   \midrule
    $32\times 32$  &  3297  &  1.32e-1 &    7.34e-1      \\

    $ 64\times 64 $  &  12737 &  1.56e-2 &    2.52e-1        \\
   
    $ 128\times 128 $ &  50049 &  2.60e-5 &    1.04e-2        \\
  
    $ 256\times 256 $ &  198401&  \textbf{1.00e-6} &    \textbf{2.07e-3}        \\
    \bottomrule
    \end{tabular}}
    \vskip -0.1in
    \label{Tabel1}
\end{table}

Figure \ref{fig2} shows a visualized comparison of the prediction performance between TT-PINNs and PINNs. As can be seen, the PINN with  $n_{\boldsymbol{\theta}}=12737$ model parameters, which corresponds to 64 neurons per layer, produces the worst prediction among the 4 models. Meanwhile, the proposed TT-PINN with only $ n_{\boldsymbol{\theta}}=3713 $ parameters, which corresponds to compressing a $256\times 256$ weight matrix by $100\times$ in the training, achieves a significantly improved prediction. Also, the TT-PINN with $ n_{\boldsymbol{\theta}}=3713 $ parameters yields an equally accurate prediction as the PINN with $ 50049 $ model parameters. These results show that, by approximating a more complicated neural network with the low rank structure (i.e., TT-cores), our proposed TT-PINNs are capable of, in some level, preserving the expressive power of a larger PINN. This will greatly reduce the requirement of hardware resources in edge computing.

\begin{table}[t]
    \caption{The performance of TT-PINNs in solving Helmholtz equation for different model sizes. Here $\widehat{\mat{W}}_{h}$ represents the weight matrix approximated by the TT-cores in each tensorized hidden layer. $n_{\boldsymbol{\theta}}$ is the total number of parameters in the TT-PINN. The mean squared errors and relative $\ell_2$ errors are reported.}
    \vskip 0.15in
    \centering
    \scalebox{0.9}{
    \begin{tabular}{ccccc}
    \toprule  
    $\widehat{\mat{W}}_{h}$ & Compression  & $n_{\boldsymbol{\theta}}$ & MSE & Rel. $\ell_{2} $ error \\
    \midrule
    $ 128 \times 128 $ & 40$\times$  &  2169  &  2.42e-4 &    3.14e-2      \\
 
    $ 128 \times 128 $ & 20$\times$  &  3597 &  3.08e-4 &    3.55e-2        \\
    
    $ 256 \times 256 $ & 100$\times$ &  3713 &  2.25e-4 &    3.03e-2        \\

    $ 256 \times 256 $ & 40$\times$ &  6593 &  1.50e-5 &    7.75e-3        \\

    $ 256 \times 256 $ & 20$\times$ &  12449 &  \textbf{4.00e-6} &    \textbf{4.26e-3}        \\
    \bottomrule
    \end{tabular}}
    \vskip -0.1in
    \label{Tabel2}
\end{table}


\section{Conclusion and Discussions}
In this paper, we have proposed an end-to-end compressed architecture for training PINNs with less computing resources. It is the first time that a low-rank structure is applied to achieve memory efficiency while maintaining satisfactory performance in training PINNs. This work is a promising solution for training PINNs on edge devices.

This work, however, is still at the early stage thus very limited. Firstly, the PDE we considered in this work is relatively simple and does not have stiffness issue that frequently occurs in many engineering problems. Secondly, the current network size we have considered is still relatively small, the performance of TT-PINN needs to be demonstrated on larger PINNs. Finally, deploying this framework on edge computing platforms (e.g., embedded GPU or FPGA) requires further algorithm/hardware co-design.

\section*{Acknowledgement}
This work was supported by NSF \# 1817037 and NSF \# 2107321.

\bibliography{example_paper}

\begin{thebibliography}{16}
\providecommand{\natexlab}[1]{#1}
\providecommand{\url}[1]{\texttt{#1}}
\expandafter\ifx\csname urlstyle\endcsname\relax
  \providecommand{\doi}[1]{doi: #1}\else
  \providecommand{\doi}{doi: \begingroup \urlstyle{rm}\Url}\fi

\bibitem[Bansal \& Tomlin(2021)Bansal and Tomlin]{bansal2021deepreach}
Bansal, S. and Tomlin, C.~J.
\newblock Deepreach: A deep learning approach to high-dimensional reachability.
\newblock In \emph{2021 IEEE International Conference on Robotics and
  Automation (ICRA)}, pp.\  1817--1824. IEEE, 2021.

\bibitem[Baydin et~al.(2018)Baydin, Pearlmutter, Radul, and
  Siskind]{baydin2018automatic}
Baydin, A.~G., Pearlmutter, B.~A., Radul, A.~A., and Siskind, J.~M.
\newblock Automatic differentiation in machine learning: a survey.
\newblock \emph{Journal of Marchine Learning Research}, 18:\penalty0 1--43,
  2018.

\bibitem[Cichocki(2014)]{cichocki2014tensor}
Cichocki, A.
\newblock Tensor networks for big data analytics and large-scale optimization
  problems.
\newblock \emph{arXiv preprint arXiv:1407.3124}, 2014.

\bibitem[Glorot \& Bengio(2010)Glorot and Bengio]{glorot2010understanding}
Glorot, X. and Bengio, Y.
\newblock Understanding the difficulty of training deep feedforward neural
  networks.
\newblock In \emph{Proceedings of the thirteenth international conference on
  artificial intelligence and statistics}, pp.\  249--256. JMLR Workshop and
  Conference Proceedings, 2010.

\bibitem[Hawkins \& Zhang(2021)Hawkins and Zhang]{hawkins2021bayesian}
Hawkins, C. and Zhang, Z.
\newblock Bayesian tensorized neural networks with automatic rank selection.
\newblock \emph{Neurocomputing}, 453:\penalty0 172--180, 2021.

\bibitem[Hawkins et~al.(2022)Hawkins, Liu, and Zhang]{hawkins2022towards}
Hawkins, C., Liu, X., and Zhang, Z.
\newblock Towards compact neural networks via end-to-end training: A {Bayesian}
  tensor approach with automatic rank determination.
\newblock \emph{SIAM Journal on Mathematics of Data Science}, 4\penalty0
  (1):\penalty0 46--71, 2022.

\bibitem[Kingma \& Ba(2014)Kingma and Ba]{kingma2014adam}
Kingma, D.~P. and Ba, J.
\newblock Adam: A method for stochastic optimization.
\newblock \emph{arXiv preprint arXiv:1412.6980}, 2014.

\bibitem[Liu \& Wang(2019)Liu and Wang]{liu2019multi}
Liu, D. and Wang, Y.
\newblock Multi-fidelity physics-constrained neural network and its application
  in materials modeling.
\newblock \emph{Journal of Mechanical Design}, 141\penalty0 (12), 2019.

\bibitem[Lu et~al.(2021)Lu, Pestourie, Yao, Wang, Verdugo, and
  Johnson]{lu2021physics}
Lu, L., Pestourie, R., Yao, W., Wang, Z., Verdugo, F., and Johnson, S.~G.
\newblock Physics-informed neural networks with hard constraints for inverse
  design.
\newblock \emph{SIAM Journal on Scientific Computing}, 43\penalty0
  (6):\penalty0 B1105--B1132, 2021.

\bibitem[Novikov et~al.(2015)Novikov, Podoprikhin, Osokin, and
  Vetrov]{novikov2015tensorizing}
Novikov, A., Podoprikhin, D., Osokin, A., and Vetrov, D.~P.
\newblock Tensorizing neural networks.
\newblock \emph{Advances in neural information processing systems}, 28, 2015.

\bibitem[Onken et~al.(2021)Onken, Nurbekyan, Li, Fung, Osher, and
  Ruthotto]{onken2021neural}
Onken, D., Nurbekyan, L., Li, X., Fung, S.~W., Osher, S., and Ruthotto, L.
\newblock A neural network approach applied to multi-agent optimal control.
\newblock In \emph{European Control Conference (ECC)}, pp.\  1036--1041, 2021.

\bibitem[Or{\'u}s(2014)]{orus2014practical}
Or{\'u}s, R.
\newblock A practical introduction to tensor networks: Matrix product states
  and projected entangled pair states.
\newblock \emph{Annals of physics}, 349:\penalty0 117--158, 2014.

\bibitem[Oseledets(2011)]{oseledets2011tensor}
Oseledets, I.~V.
\newblock Tensor-train decomposition.
\newblock \emph{SIAM Journal on Scientific Computing}, 33\penalty0
  (5):\penalty0 2295--2317, 2011.

\bibitem[Raissi et~al.(2019)Raissi, Perdikaris, and
  Karniadakis]{raissi2019physics}
Raissi, M., Perdikaris, P., and Karniadakis, G.~E.
\newblock Physics-informed neural networks: A deep learning framework for
  solving forward and inverse problems involving nonlinear partial differential
  equations.
\newblock \emph{Journal of Computational physics}, 378:\penalty0 686--707,
  2019.

\bibitem[Raissi et~al.(2020)Raissi, Yazdani, and Karniadakis]{raissi2020hidden}
Raissi, M., Yazdani, A., and Karniadakis, G.~E.
\newblock Hidden fluid mechanics: Learning velocity and pressure fields from
  flow visualizations.
\newblock \emph{Science}, 367\penalty0 (6481):\penalty0 1026--1030, 2020.

\bibitem[Stevens et~al.(2020)Stevens, Taylor, Nichols, Maccabe, Yelick, and
  Brown]{stevens2020ai}
Stevens, R., Taylor, V., Nichols, J., Maccabe, A.~B., Yelick, K., and Brown, D.
\newblock {AI} for science.
\newblock Technical report, Argonne National Lab.(ANL), Argonne, IL (United
  States), 2020.

\end{thebibliography}
\bibliographystyle{icml2022}



\end{document}